\pdfoutput=1

\documentclass[11pt]{article}

\usepackage[preprint]{acl}
\usepackage{booktabs}
\usepackage{longtable}
\usepackage{float}  

\usepackage{amsmath}    
\usepackage{amssymb}    
\usepackage{amsthm}     
\usepackage{mathtools}  
\usepackage{bm}         
\usepackage{mathrsfs}   
\usepackage{pifont} 

\usepackage[utf8]{inputenc}
\usepackage[T1]{fontenc} 
\usepackage{placeins}
\usepackage[export]{adjustbox}
\usepackage{tablefootnote}
\usepackage{algorithm}
\usepackage{adjustbox}
\usepackage{algpseudocode}
\usepackage{booktabs}       
\usepackage{xcolor}
\usepackage{multirow}
\usepackage{graphicx}
\usepackage{lipsum} 
\usepackage{times}
\usepackage{latexsym}

\usepackage[T1]{fontenc}

\usepackage[utf8]{inputenc}

\usepackage{microtype}

\usepackage{inconsolata}

\usepackage{graphicx}

\title{Guardrails in Logit Space: Safety Token Regularization for LLM Alignment Preservation}

\author{Thong Bach\\
        s224726627@deakin.edu.au }

\author{
Thong Bach\thanks{\ t.bach@deakin.edu.au} \quad
Truyen Tran
\\[0.5em]
Applied Artificial Intelligence Initiative (A2I2), Deakin University
}

\begin{document}
\maketitle

\begin{abstract}

Fine-tuning well-aligned large language models (LLMs) on new domains often degrades their safety alignment, even when using benign datasets. Existing safety alignment techniques primarily focus on pretraining, leaving fine-tuned models vulnerable to behavioral shifts. In this work, we introduce safety token regularization (STR), a lightweight method designed to preserve safety properties during fine-tuning. Our approach identifies salient tokens from rejection templates of well-aligned models and constrains their associated logits during training, preventing the loss of critical safety behaviors. Unlike reinforcement learning or preference optimization methods, STR requires minimal additional computation and seamlessly integrates with parameter-efficient fine-tuning techniques such as LoRA. Comprehensive experiments demonstrate that our approach achieves safety performance on par with state-of-the-art methods, while preserving task-specific utility and requiring minimal implementation overhead. Furthermore, we show that safety token regularization enhances training stability and overall performance beyond safety considerations alone. This work offers a practical and readily deployable strategy for continual safety alignment in fine-tuned LLMs.
\end{abstract}
\section{Introduction}
\begin{figure}[t]
    \centering
    \includegraphics[width=1.0\linewidth]{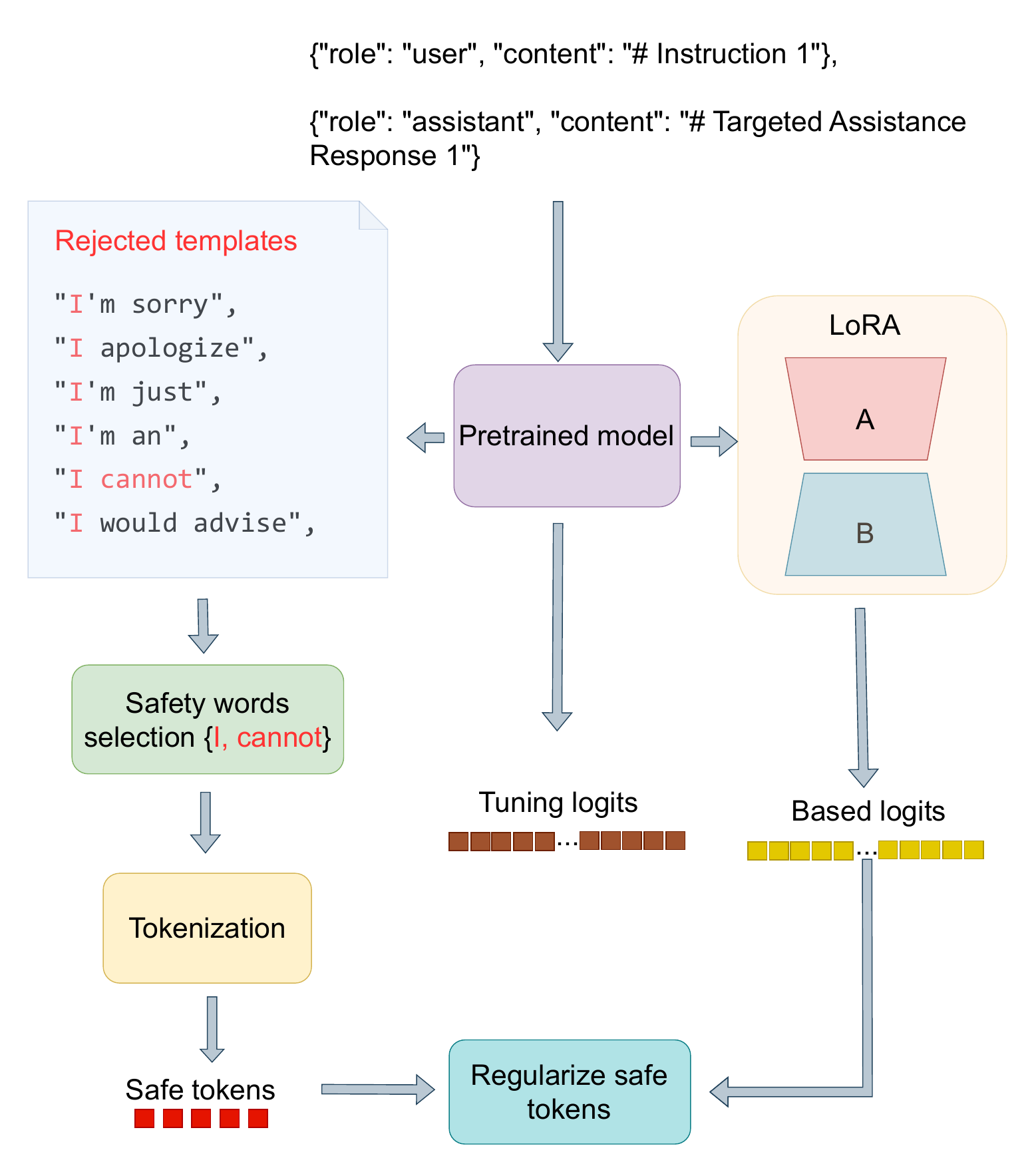} 
    \caption{\textbf{STR architecture overview.} Our method identifies “safety words” from the rejected templates of well-aligned models. These words are then tokenized to generate a set of “safe tokens”, which are used to constrain the model’s behavior in the logits space, ensuring that the fine-tuned model remains consistent with the pretrained model’s safety standards.}
    \label{fig:main_arc}
\end{figure}

The rapid integration of Large Language Models (LLMs) into real-world applications, particularly in sensitive areas like healthcare, education, and law, demands that LLMs remain safe and aligned with human values, especially when fine-tuned for specific tasks \cite{bommasani2021opportunities,thirunavukarasu2023large,hager2024evaluation,wei2022emergent}. While pre-training of frontier LLMs attempts to instill fundamental safety behaviors, it is concerning that fine-tuning, even on benign and useful datasets, can inadvertently erode this crucial pre-trained safety alignment \cite{zhao2023learning}. 


Existing approaches to safety alignment, such as Reinforcement Learning from Human Feedback \cite{bai2022training,sun2023principle} and Direct Preference Optimization \cite{rafailov2023direct,meng2025simpo}, have primarily focused on pre-training and initial alignment. While Parameter-Efficient Fine-Tuning (PEFT) techniques like LoRA \cite{hu2021lora} offer efficient adaptation, they often lack explicit mechanisms to preserve pre-trained safety during task-specific fine-tuning. Recent research has begun to address this critical gap, exploring methods to extract and maintain safety patterns during fine-tuning \cite{hu2024gradient, zhao2023learning,hsu2024safe,peng2024navigating,qi2024safety}. However, a need remains for \emph{simple, effective, and readily deployable} techniques that can robustly preserve pre-trained safety without compromising task performance.

In this work, we introduce safety token regularization (STR), a novel approach to continual safety alignment during fine-tuning. Our core insight is that pre-trained models exhibit distinct patterns in their responses to safety-critical prompts, often manifested through specific safety-indicative tokens. We propose a method to identify these tokens by analyzing the pre-trained model's behavior and then introduce a regularization term during fine-tuning that constrains the logits of these tokens. This constraint ensures that the fine-tuned model retains its sensitivity to safety signals, mirroring the pre-trained model's behavior in safety-relevant contexts. See Fig.~\ref{fig:main_arc} for model architecture.


We present extensive experimental evaluations on benchmark datasets, comparing our method against state-of-the-art safety fine-tuning techniques. Our results demonstrate that safety token regularization achieves safety performance on par with, or exceeding, current methods, while maintaining or even improving performance on target tasks. Moreover, we analyze the impact of key hyperparameters and provide insights into the mechanism by which our method effectively preserves safety.
Our findings suggest that while token-level constraints alone may be insufficient for comprehensive safety alignment, they can serve as a lightweight regularization strategy to improve safety retention in continually fine-tuned LLMs.

Thus our approach offers several key advantages: it is conceptually simple and easily integrated into existing PEFT frameworks, demonstrably preserves pre-trained safety alignment, maintains competitive task-specific performance, and, surprisingly, can even enhance post-training stability and overall model utility. Our main contributions are:

>We introduce safety token regularization (STR), a token-level safety preservation method that enhances fine-tuning without the need for additional preference modeling or adversarial training.

>We conduct extensive experiments on multiple benchmarks, demonstrating that STR reduces harmful response rates in fine-tuned models while maintaining competitive utility.

>Further experiments show that, beyond improving safety, token regularization enhances the stability and overall performance of PEFT models.

\section{Related works}
\paragraph{Parameter-efficient fine-tuning in LLMs.}
As models scale from millions to trillions of parameters, full fine-tuning becomes increasingly challenging and intractable for most researchers. Consequently, efficient fine-tuning methods have become crucial. Several approaches have emerged in this space, including prefix-tuning \cite{li2021prefix,jia2022visual,liu2024improved,zhou2022conditional}, representation editing \cite{wu2025reft,li2024inference,hernandez2023inspecting,xu2025uncovering}, and Low-Rank Adaptation (LoRA) \cite{hu2021lora,dettmers2023qlora}. Among these, LoRA has gained widespread adoption due to its memory efficiency and its ability to achieve performance comparable to full fine-tuning across various conditions. Recent research has explored ways to enhance LoRA and address its limitations from different aspects. Work by \cite{liu2024dora} decomposed pretrained weights into magnitude and direction components, using LoRA to fine-tune only the directions, further reducing trainable parameters. Besides, \cite{dettmers2023qlora} introduced a quantization approach for LoRA, optimizing both training time and memory efficiency. On the other hand, \cite{meng2025pissa} proposed initialization techniques based on singular values to accelerate LoRA's convergence. In this work, we adopt LoRA as our baseline and incorporate safety regularization within the PEFT framework, integrating it into both the training and evaluation phases.

\paragraph{Safety alignment.}
Aligning LLMs to follow human rules is crucial for their deployment in real applications. RLHF, the pioneering approach in this direction, utilizes reinforcement learning and human preferences to teach models to follow human values \cite{bai2022training,dai2023safe}. DPO \cite{rafailov2023direct} further enhances RLHF by introducing reward-free optimization mechanisms that reduce alignment costs and ensure procedural stability. Subsequently, extensive research has emerged aimed at enhancing the alignment process through cost reduction \cite{meng2025simpo,huang2024one,ji2025aligner}, robustness improvement \cite{bach2026curvatureaware,bach2026rethinking,zheng2023improving}, and increased sample efficiency \cite{sun2023principle,zhou2024lima}. Given their impressive results and successful alignment with human values, these approaches were primarily adopted during pre-training rather than fine-tuning. However, this creates a new safety risk when fine-tuning these models on new data, as first demonstrated in \cite{qi2023fine}. Consequently, ensuring safety during fine-tuning has gained increased attention \cite{hsu2024safe,li2025salora,zhao2023learning,hu2024gradient,he2024s}. Research by \cite{hu2024gradient} filters high-risk samples based on gradient norms, while \cite{zhao2023learning} demonstrates that models tend to forget unsafe samples more readily than benign ones. Additionally, \cite{huang2024vaccine} proposes robust training techniques to prevent harmful behavior. From a PEFT perspective, \cite{hsu2024safe} and \cite{li2025salora} extract safety patterns from pretrained models and introduce mechanisms to preserve these beneficial patterns during the fine-tuning process.
\section{Method}
\label{sec:method}

In this section, we describe our light-weight \emph{safety-preserving} fine-tuning approach, which can be seamlessly integrated into parameter-efficient fine-tuning (PEFT) frameworks. Our method enforces logit constraints on a predefined set of ``safety tokens'', ensuring that fine-tuned models preserve their pretrained refusal behavior while maintaining task-specific performance. 

\subsection{Preliminaries}

We consider a fine-tuning dataset 
$\mathcal{D}$ consisting of 
\(N\) sequence pairs,
$
\mathcal{D} 
= \{(x^{(n)}_{1:T_n}, y^{(n)}_{1:T_n})\}_{n=1}^N
$, 
where \(x_{1:T_n}^{(n)}\) is the input tokens (prompt or partially observed sequence); 
\(y_{1:T_n}^{(n)}\) is the target tokens to be predicted autoregressively; 
and  \(T_n\) is the sequence length of the \(n\)-th example.
This dataset $\mathcal{D}$ is used to fine-tune the pre-trained language model using our proposed safety token regularization approach.

\subsection{Identifying safety tokens}
\label{subsec:safety_tokens}
We define a set of \emph{safety tokens}, \(\{t_k\}_{k=1}^K\) as those words or subwords associated with disallowed content, hate speech, or other high-risk categories.
One way to identify these tokens is through a vocabulary-based analysis of harmful expressions. However, these do not necessarily refect how the base models handle the tokens.

Our work adopts a more direct approach by analyzing how aligned models respond to harmful queries. That is, we identify safety-indicative tokens – those that are strongly associated with the pre-trained model's safety-oriented behavior. Our hypothesis, illustrated conceptually in Fig.~\ref{fig:main_arc}, is that well-aligned pre-trained models exhibit consistent patterns in their responses when confronted with potentially harmful or inappropriate queries. These patterns often manifest in the form of specific rejection templates, which aligned models employ to gracefully refuse or deflect such requests.

Based on this observation, we selected common words from these rejection templates as safety words. To demonstrate our method in experiments, we used three common words: \{`I,' `cannot,' and `can't'\}. These words were then tokenized to determine the corresponding safety tokens. Our experimental results in the next section will demonstrate that preserving the logits of these safety tokens during fine-tuning maintains the model's safety behavior while stabilizing the learning process.

\subsection{PEFT loss}
\label{subsec:single_example_loss}
For clarity, consider a training instance \(\bigl(x^{(n)}_{1:T_n}, y^{(n)}_{1:T_n}\bigr)\). we define a composite loss function comprising two key components: Autoregressive cross-entropy and Safety token regularization.

\paragraph{Autoregressive cross-entropy.}
We adopt the standard cross-entropy for autoregressive language modeling:
\begin{equation}
    \label{eq:ce_loss_single}
    \mathcal{L}_\text{CE}^{(n)} 
    = - \frac{1}{T_n} \sum_{t=1}^{T_n}
    \log P\bigl(y^{(n)}_t \mid x^{(n)}_{1:t-1}; \Theta_\text{PEFT}\bigr),
\end{equation}
where \(\Theta_\text{PEFT}\) denotes the trainable PEFT parameters (e.g., using LoRA of \cite{hu2021lora}), while the base model parameters \(\Theta_\text{base}\) remain frozen. This loss encourages predictions to match the ground-truth tokens. As PEFT does not explicitly preserve safety alignment, we introduce safety token regularization, constraining the model’s logits on critical safety tokens.

\paragraph{Safety token regularization.}
To preserve safe behavior the pre-trained model during fine-tuning, we introduce a safety token regularization (STR) term. This term penalizes deviations in the logits of pre-defined ``safety token'' between the base pre-trained model and the PEFT-adapted model.  The logits are particulary informative about the corresponding tokens in relative preferences to other tokens in the vocabulary in the same context, indicative of the confidence in generating the tokens as well as semantic relationships between tokens.  

Let 
$\ell_{t,k}^\text{base}$ and $\quad \ell_{t,k}^\text{PEFT}$ 
denote the logits of the \(k\)-th safety token at time step \(t\) under the base model (\(\Theta_\text{base}\)) and the PEFT-adapted model (\(\Theta_\text{PEFT}\)), respectively. We deviations in the logits can be captured via a square loss:
\begin{equation}
    \label{eq:frobenius_loss_single}
    \mathcal{L}_\text{F}^{(n)}
    = \frac{1}{T_n} \sum_{t=1}^{T_n} \sum_{k=1}^K
      \left(\ell_{t,k}^\text{base} - \ell_{t,k}^\text{PEFT}\right)^2.
\end{equation}
Minimizing this term constrains the PEFT-updated logits for safety tokens to remain close to those of the base model, effectively preserving its original safety-oriented behavior in the logit space. 

\paragraph{Combined loss.}
We form the total loss for the \(n\)-th example by:
\begin{equation}
    \label{eq:single_example_total_loss}
    \mathcal{L}^{(n)} = \mathcal{L}_\text{CE}^{(n)} + \lambda \,\mathcal{L}_\text{F}^{(n)},
\end{equation}
where \(\lambda > 0\) controls the trade-off between modeling accuracy and safety-token consistency.

The loss for the full fine-tuning dataset \(\mathcal{D}\) of \(N\) instances is simply the averaging of the instance losses. See Algorithm~\ref{alg:training_procedure_short} for a pseudocode for the entire fine-tuning process.

\begin{algorithm}[t]
\caption{Safety-Aware PEFT Fine-Tuning}
\label{alg:training_procedure_short}
\begin{algorithmic}[1]
\Require 
  \(\Theta_{\text{base}}\): Frozen pretrained model \quad
  \(\Theta_{\text{PEFT}}\): Trainable PEFT params \\
  \(\mathcal{D}\): Training set of sequences \quad
  \(\{t_k\}_{k=1}^K\): Safety tokens \quad
  \(\lambda\): Frobenius norm weight \quad
  \(\eta\): Learning rate
\Ensure \(\Theta_{\text{PEFT}}\) (updated)

\State Initialize \(\Theta_{\text{PEFT}}\); Freeze \(\Theta_{\text{base}}\)
\For{each training iteration}
  \State Sample batch \(\{(x_{1:T}, y_{1:T})\}\) from \(\mathcal{D}\)
  \State \textbf{Base Model} Get \(\ell_{t,k}^{\text{base}}\) for safety tokens
  \State \textbf{PEFT Forward:} Get \(\ell_{t,k}^{\text{PEFT}}\) and compute
    \[
      P\bigl(y_t \!\mid\! x_{1:t-1}; \Theta_{\text{PEFT}}\bigr)
    \]
  \State \textbf{Losses:} Compute $\mathcal{L}_\text{CE}$, $\mathcal{L}_\text{F}$ and
    \[
      \mathcal{L} = \mathcal{L}_\text{CE} + \lambda \,\mathcal{L}_\text{F}
    \]
  \State \textbf{Update \(\Theta_{\text{PEFT}}\):} 
    \[
      \Theta_{\text{PEFT}} \leftarrow 
      \Theta_{\text{PEFT}} - \eta\,\nabla_{\Theta_{\text{PEFT}}}\mathcal{L}
    \]
\EndFor

\end{algorithmic}

\end{algorithm}





\section{Experiments}

\begin{table*}[t]
\centering
\begin{adjustbox}{width=0.7\textwidth}
\begin{tabular}{l cc cc}
\toprule
& \multicolumn{2}{c}{Llama-2-chat-7B} & \multicolumn{2}{c}{Llama-3.1-Instruct-8B} \\
\cmidrule(lr){2-3}\cmidrule(lr){4-5}
\textbf{Before Fine-tuning} & \multicolumn{2}{c}{0.0\%} & \multicolumn{2}{c}{1.4\%} \\
\textbf{Rank for PEFT Training} & 16 & 32 & 16 & 32 \\
\midrule
\textbf{Base PEFT} \\
\quad LORA & 23.7\% & 31.7\% & 13.8\% & 14.5\% \\
\quad PiSSA & 31.7\% & 35.7\% & 13.8\% & 14.5\% \\
\quad DORA & 23.7\% & 25.3\% & 10.1\% & 9.4\% \\
\midrule
\textbf{LoRA w.\ post-hoc alignment} \\
\quad LORA w.\ IA & 13.5\% & 23.7\% & 7.7\% & 5.8\% \\
\quad LORA w.\ Vac & 20.2\% & 25.3\% & 41.1\% & 38.3\% \\
\quad Safe LoRA & 15.7\% & 14.5\% & 8.5\% & 6.7\% \\
\quad SaLoRA & 3.5\% & 4.4\% & \textbf{2.9\%} & \textbf{1.4\%} \\
\midrule
\quad \textbf{Ours (I)} & \textbf{2.9\%} & \textbf{0.0\%} & 3.1\% & 1.5\% \\
\quad \textbf{Ours (cannot)} & 3.4\%& 0.6\% & 3.5\% & 1.5\% \\
\bottomrule
\end{tabular}
\end{adjustbox}
\caption{\textbf{Harmful Response Rate (HRR) on Aplaca dataset.}
Overall, our method achieves results on par with current state-of-the-art methods across different settings. Notably, under the LLaMA-2 setup, our approach outperforms competing methods, and matches the safety performance of pretrained models on rank 32.}
\label{tab:alpaca_result}
\end{table*}

\begin{table}[h]
    \centering
\renewcommand{\arraystretch}{1.2} 
    \setlength{\tabcolsep}{10pt} 
    \begin{adjustbox}{width=0.48\textwidth}
    \begin{tabular}{l|l|c}
        \toprule
        \textbf{Models} & \textbf{Method} & \textbf{HRR(↓)} \\
        \midrule
        \multirow{5}{*}{LM2-Chat} 
        & Pretrained & 0.0\% \\
        & LoRA & 62.3\% \\
        & \textbf{Ours (I)} & 0.0\% \\
        & \textbf{Ours (cannot)} & 0.0\% \\
        \midrule
        \multirow{5}{*}{LM3-Instruct} 
        & Pretrained & 12.7\% \\
        & LoRA & 47.3\% \\
        & \textbf{Ours (I)} & 13.5\% \\
        & \textbf{Ours (cannot)} & 7.7\% \\
        & \textbf{Ours (I cannot)} & 4.0\% \\
        \bottomrule
    \end{tabular}
    \end{adjustbox}
    \caption{\textbf{Harmful Response Rate (HRR) on PureBad dataset.}
    Our approach preserves safety behavior in LLaMA-2 and even surpasses the safety performance of the pretrained LLaMA-3 models. These findings suggest that our method can learn beyond the baseline safety provided by the pretrained model.
    }
    \label{tab:pure_bad}
\end{table}


\subsection{Experimental setup}

To comprehensively evaluate the effectiveness of our proposed safety token regularization method, we conducted experiments across a diverse set of benchmarks, focusing on both \textbf{safety} and \textbf{utility} aspects of fine-tuned LLMs. We utilized two prominent LLM architectures: LLaMA-2-7b-chat and LLaMA-3-8b-instruct, representing models of varying scales and pre-training paradigms. For parameter-efficient fine-tuning, we employed LoRA, a widely adopted technique known for its efficiency and performance.

The regularization weight $\lambda$ in Eq.~(\ref{eq:single_example_total_loss}) was selected via validation.
Generally, we use larger \(\lambda\) values (more constraints) for high-risk datasets and smaller values for benign data. Notably, we found that commonsense reasoning tasks \cite{hu2023llm} require a small \(\lambda\) value ($0.1$ in our setting), aligning with recent findings that improved reasoning capacity correlates with safer model behavior \cite{guan2024deliberative}.

\begin{table*}[h]
\centering
\renewcommand{\arraystretch}{1.2} 
\setlength{\tabcolsep}{8pt} 
\begin{adjustbox}{width=1.0\textwidth}
\begin{tabular}{l|ccccc ccc|cc}
    \hline
    \textbf{Method} & \textbf{BoolQ} & \textbf{PIQA} & \textbf{SIQA} & \textbf{HellaSwag} & \textbf{WinoGrande} & \textbf{ARC-e} & \textbf{ARC-c} & \textbf{OBQA} & \textbf{Avg. Acc} & \textbf{ASR} \\
    \hline
    \multicolumn{11}{c}{r=16} \\
    \hline
    LoRA  & \textbf{65.7\%} & 75.4\% & 70.1\% & \textbf{53.9\%} & 66.0\% & 76.3\% & 61.2\% & 66.0\% & 66.8\% & \textbf{0.19\%} \\
    DoRA  & 64.8\% & 75.8\% & \textbf{74.2\%} & 42.0\% & 64.9\% & \textbf{82.5\%} & \textbf{66.0\%} & \textbf{76.0\%} & \textbf{68.3\%} & \textbf{0.19\%} \\
    \textbf{Ours} & 65.2\% & \textbf{77.3\%} & 67.1\% & 46.8\% & \textbf{67.9\%} & 78.4\% & 64.3\% & 75.5 & 67.8\% & \textbf{0.19\%} \\
    \hline
    \multicolumn{11}{c}{r=32} \\
    \hline
    LoRA  & 62.4\% & \textbf{75.7\%} & 42.4\% & 27.7\% & 63.7\% & 50.0\% & 41.0\% & 43.8\% & 50.8\% & 0.19\% \\
    DoRA  & \textbf{63.1\%} & 62.6\% & 31.4\% & 31.1\% & 62.1\% & 43.9\% & 32.8\% & 42.3\% & 46.2\% & 0.19\% \\
    \textbf{Ours} & 58.5 & 71.2\% & \textbf{71.0\%} & \textbf{45.4\%} & \textbf{67.2\%} & \textbf{82.3\%} & \textbf{67.5\%} & \textbf{76.7\%} & \textbf{67.5\%} & \textbf{0.0\%} \\
    \hline
    \multicolumn{11}{c}{r=64} \\
    \hline
    LoRA  & 61.9\% & 58.9\% & 49.7\% & 39.5\% & 56.3\% & 64.0\% & 51.5\% & 56.5\% & 54.8\% & 0.19\% \\
    DoRA  & \textbf{63.1\%} & 73.6\% & 48.3\% & 39.2\% & 60.5\% & 36.9\% & 31.3\% & 36.0\% & 48.6\% & \textbf{0.0\%} \\
    \textbf{Ours} & 59.7\% & \textbf{77.4\%} & \textbf{73.2\%} & \textbf{56.3\%} & \textbf{72.2\%} & \textbf{82.5\%} & \textbf{68.2\%} & \textbf{76.8\%} & \textbf{70.8\%} & \textbf{0.0\%} \\
    \hline
\end{tabular}
\end{adjustbox}
\caption{\textbf{Accuracy and ASR Comparison.} The table summarizes the performance of LoRA, DoRA, and our method when fine-tuning the Commonsense-15K dataset and evaluating across five commonsense reasoning tasks in terms of accuracy and Attack Success Rate (ASR). At a low rank (r=16), all methods exhibit stable performance, with our approach trailing DoRA slightly. However, as the rank increases, LoRA and DoRA become unstable and degrade significantly, whereas our method remains robust and outperforms both baselines by a substantial margin. From a safety perspective, all methods demonstrate high safety capacity, likely due to the focus on reasoning data during fine-tuning.}
\label{tab:commonsense}
\end{table*}

\begin{table*}[h]
    \centering
    \renewcommand{\arraystretch}{1.2} 
    \setlength{\tabcolsep}{8pt} 
    \begin{adjustbox}{width=0.75\textwidth}
    \begin{tabular}{l|ccccc|c}
        \hline
        \textbf{Method} & \textbf{BoolQ} & \textbf{SIQA} & \textbf{PIQA} & \textbf{WinoGrande} & \textbf{ARC-c} & \textbf{Avg. Accuracy} \\
        \hline
        \multicolumn{7}{c}{r=32} \\
        \hline
        LoRA  & 71.2\%  & 80.2\%  & 79.1\%  & 70.7\%  & 65.3\%  & 73.3\%  \\
        DoRA  & 62.8\%  & 78.1\%  & 81.7\%  & 82.9\%  & 61.9\%  & 73.5\%  \\
        Ours  & 68.3\%  & 78.9\%  & 81.2\%  & 80.8\%  & 64.2\%  & \textbf{74.6\%}  \\
        \hline
        \multicolumn{7}{c}{r=64} \\
        \hline
        LoRA  & 56.0\%  & 76.9\%  & 80.4\%  & 81.8\%  & 63.7\%  & 71.8\%  \\
        DoRA  & 62.2\%  & 7.5\%   & 79.3\%  & 77.1\%  & 30.3\%  & 51.3\%  \\
        Ours  & 62.2\%  & 78.6\%  & 76.4\%  & 81.4\%  & 61.2\%  & \textbf{72.0\%}  \\
        \hline
        \multicolumn{7}{c}{r=128} \\
        \hline
        LoRA  & 69.7\%  & 48.1\%  & 81.7\%  & 82.7\%  & 64.2\%  & 69.3\%  \\
        DoRA  & 67.2\%  & 73.9\%  & 47.0\%  & 6.2\%   & 23.0\%  & 43.5\%  \\
        Ours  & 62.2\%  & 74.6\%  & 80.8\%  & 81.6\%  & 64.8\%  & \textbf{72.8\%}  \\
        \hline
        \multicolumn{7}{c}{r=256} \\
        \hline
        LoRA  & 62.1\%  & 80.2\%  & 82.5\%  & 83.3\%  & 67.0\%  & 75.0\%  \\
        DoRA  & 5.7\%   & 28.4\%  & 33.8\%  & 14.7\%  & 22.9\%  & 21.1\%  \\
        Ours  & 69.6\%  & 79.4\%  & 83.0\%  & 83.8\%  & 66.3\%  & \textbf{76.4\%}  \\
        \hline
        
    \end{tabular}
    \end{adjustbox}
    \caption{\textbf{Continual Learning Performance.}The Table presents the accuracy of LoRA, DoRA, and our method across various datasets under continual learning conditions. Our method achieves state-of-the-art performance in all settings. Notably, when the rank is increased, both LoRA and DoRA exhibit instability—mirroring observations from earlier experiments—whereas our method remains stable at higher ranks, further widening the performance gap relative to other PEFT approaches.}
    \label{tab:continual}
\end{table*}

\subsection{Evaluation datasets and metrics}

Our evaluation encompassed three distinct experimental settings, each designed to assess different facets of \textbf{safety} and \textbf{utility}:

\textbf{Alpaca dataset for safety evaluation when fine-tuning on benign data}: This instruction-following dataset \cite{taori2023alpaca}, containing over 50,000 samples, is widely used to assess the safety preservation in a more general fine-tuning scenario. Following \cite{li2025salora}, we trained models for one epoch, reserving 200 samples for evaluation. 

\textbf{PureBad dataset for harmfulness evaluation}:
This dataset--constructed from 100 highly harmful prompts extracted from \cite{qi2023fine}--was used to
evaluate the method's ability to preserve safety behavior under adversarial conditions. 
Fine-tuning on this dataset serves as a stress test, revealing the model's robustness against safety degradation when exposed to exclusively harmful content.
We applied LoRA with rank $8$ to both LLaMA-2 and LLaMA-3 models using a learning rate of $0.0005$. The fine-tuning ran for 5 epochs on LLaMA-2, while LLaMA-3 needed longer training of 30 epochs (we extended the training time because LLaMA-3 stayed safe after 5 epochs, and we wanted to test its behavioral limits, similar to what \cite{li2025salora} observed).

\textbf{Commonsense-15k dataset for utility evaluation}: To assess the impact on utility, particularly in reasoning capabilities, we fine-tuned on the Commonsense-15k dataset \cite{liu2024dora}. We evaluated performance on eight diverse commonsense reasoning benchmarks: BoolQ, SIQA, PIQA, HellaSwag, WinoGrande, ARC-e, ARC-c, and OBQA.



\textbf{Evaluation metrics.} For safety, we report the Harmful Response Rate (HRR), calculated as the percentage of generated responses flagged as harmful by an automated safety classifier \cite{dubey2024llama3herdmodels}, following the evaluation protocol of \cite{li2025salora}. 
Another key metric for safety is the Attack Success Rate (ASR), which is evaluated using keyword matching, following \cite{qi2023fine}. For utility, we report Average Accuracy across these tasks as the primary utility metric.

\subsection{Safety on Alpaca dataset}
We compared our method against current state-of-the-art approaches presented in \cite{li2025salora}, including PEFT baselines—LoRA \cite{hu2021lora}, DoRA \cite{liu2024dora}, and PiSSA \cite{meng2025pissa}—and LoRA combined with post-hoc alignment methods: InferAligner (IA) \cite{wang2024inferaligner}, and Vaccine (Vac) \cite{huang2024vaccine}.
As shown in Table~\ref{tab:alpaca_result}, our method achieves competitive results compared to current state-of-the-art approaches.
On LLaMA-2, our models establish new state-of-the-art performance, matching pretrained model safety levels at rank $r=32$. For LLaMA-3, our approach achieves comparable results to existing methods. Consistent with previous findings in \cite{li2025salora}, larger LLaMA-3 models demonstrate greater robustness during fine-tuning compared to their smaller counterparts.

\subsection{Safety on PureBad dataset}
Table~\ref{tab:pure_bad} presents the Harmful Response Rate (HRR) on the PureBad dataset for LLaMA-2-7b-chat and LLaMA-3-8b-instruct models fine-tuned using LoRA and our safety token regularization method, across LoRA ranks of $16$ and $32$. As shown, standard LoRA fine-tuning significantly degrades safety, resulting in HRRs of $62.3\%$ and $47.3\%$ for LLaMA-2-7b-chat and LLaMA-3-8b-instruct, respectively. In stark contrast, our safety token regularization method effectively mitigates this safety degradation, achieving HRRs of $0.0\%$ for LLaMA-2-7b-chat and $13.5\%$ for LLaMA-3-8b-instruct. Crucially, for LLaMA-2-7b-chat, \emph{our method restores safety performance to the level of the pre-trained model ($0.0\%$ HRR) and even surpasses the pre-trained safety of LLaMA-3-8b-instruct ($12.7\% HRR$)}, demonstrating its ability to not only preserve but potentially enhance pre-existing safety characteristics. These results strongly indicate the effectiveness of safety token regularization in maintaining safety robustness, even when fine-tuning on exclusively harmful data.


\subsection{Utility on Commonsense-15k dataset}
\begin{figure*}[h]
    \centering
    \includegraphics[width=0.85\linewidth]{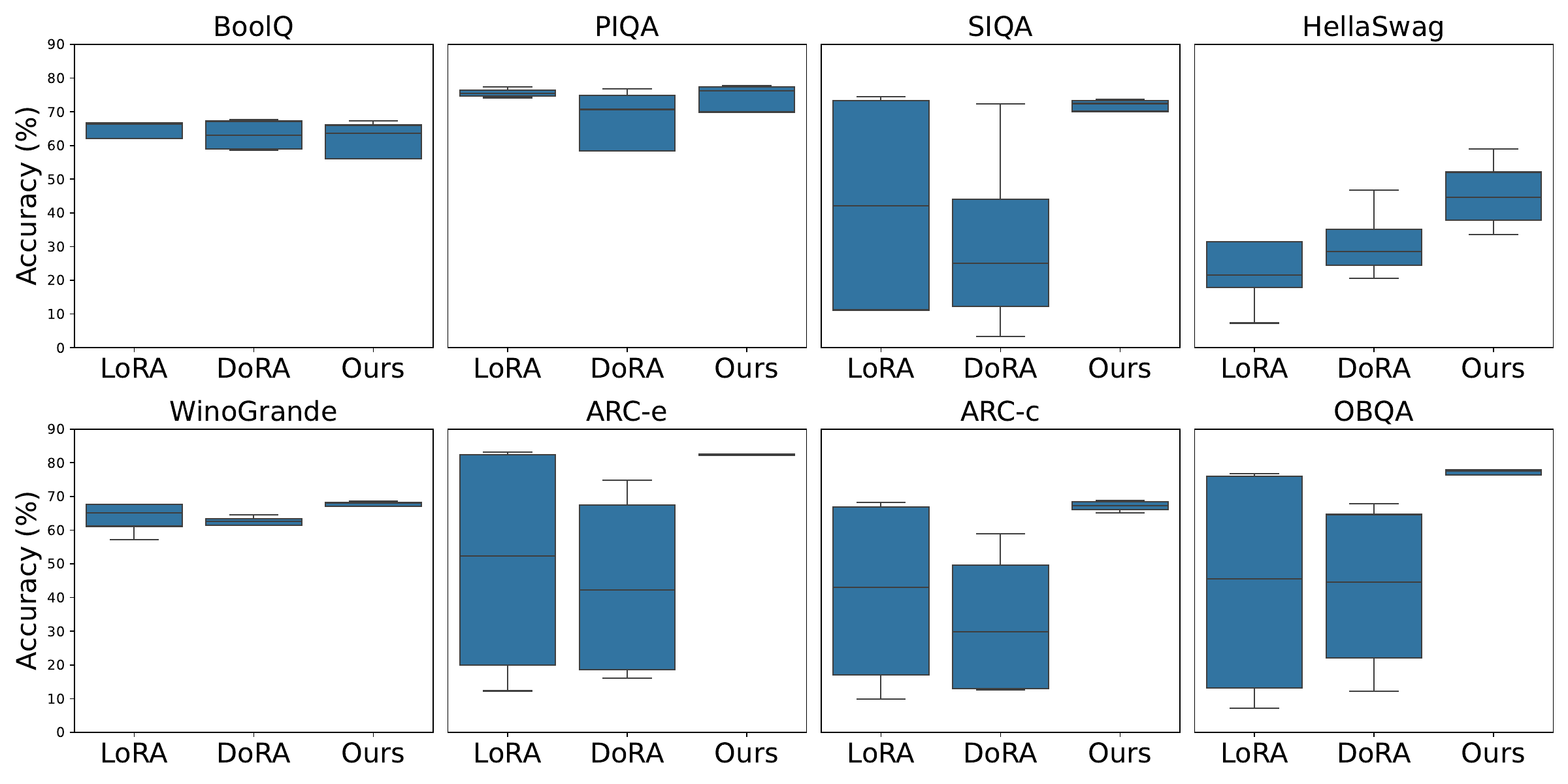} 
    \caption{
The figure presents a box plot comparison of LoRA, DoRA, and our method on eight commonsense datasets at rank 32 when tuning on the Commonsense-15k dataset. Overall, our approach consistently exceeds the performance of both LoRA and DoRA while exhibiting greater stability across multiple runs. These findings underscore the robustness and reliability of our method under varied conditions.}
    \label{fig:stable}
\end{figure*}

We followed the experimental settings from \cite{liu2024dora} to compare our method's utility against common PEFT methods like LoRA and DoRA. 
Table~\ref{tab:commonsense} and Fig.~\ref{fig:stable} present the Average Accuracy and accuracy variations across individual tasks, respectively, for LoRA, DoRA, and our method fine-tuned on the Commonsense-15k dataset. At lower LoRA ranks (e.g., $r=16$), all methods exhibit comparable Average Accuracy, suggesting that safety token regularization does not significantly hinder initial learning capacity. However, as the LoRA rank increases ($r=32$, $r=64$), standard LoRA and DoRA demonstrate increasing instability and performance degradation, as evidenced by the widening confidence intervals in Fig.~\ref{fig:stable} and decreasing Average Accuracy in Table~\ref{tab:commonsense}. In stark contrast, our safety token regularization method maintains consistent and robust performance even at higher ranks, achieving the highest Average Accuracy ($70.8\%$ at $r=64$) and exhibiting significantly reduced accuracy variance across tasks (Fig.~\ref{fig:stable}). This enhanced stability and sustained utility at higher ranks suggest that safety token regularization not only preserves safety but may also contribute to more robust and reliable fine-tuning, particularly when increasing model capacity for complex tasks.

\subsection{Continual learning with safe tokens}
In addition to the standard evaluation settings, we also assess \textit{safety tokens} in \textbf{continual learning} scenarios. Specifically, we conduct experiments on five commonsense reasoning tasks—BoolQ, PIQA, SIQA, WinoGrande, and ARC-c—using the LLaMA-2-7b-chat model. In this setup, each task is learned sequentially without access to previous or future tasks’ data, and we evaluate performance on models trained across all tasks in order. We compare our approach with LoRA and DoRA under the same experimental conditions described by \cite{liu2024dora}, except that we reduce the training epochs from three to one. As shown in Table~\ref{tab:continual}, our method consistently achieves higher average accuracy across all rank settings than LoRA and DoRA. In addition to the overall performance gain, our method demonstrates more stable performance—particularly at higher ranks—mirroring the training behavior observed on the commonsense-15k dataset.

\subsection{Safety of random tokens}
\begin{figure}[t]
    \centering
    \includegraphics[width=0.9\linewidth]{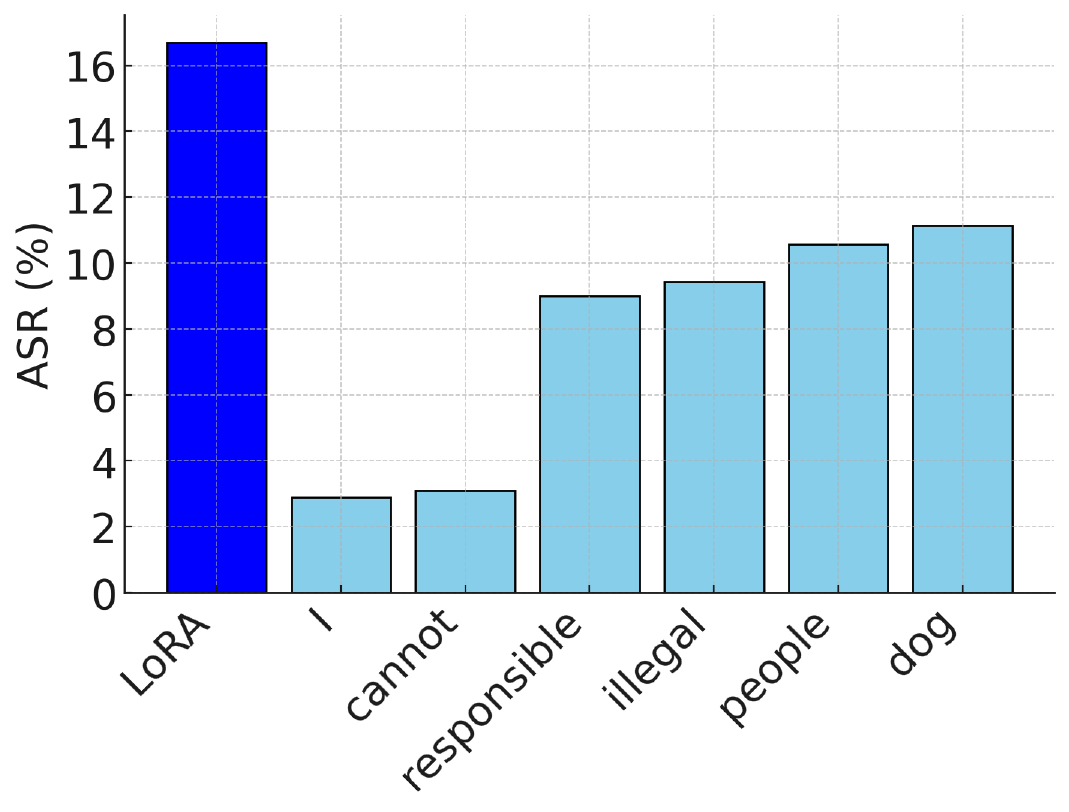} 
    \caption{\textbf{Safe and Random Tokens Performance.}
We compare the effects of “safe” versus “random” tokens on the Alpaca dataset trained using the LLaMA-2 model. Despite being randomly selected, these tokens still enhance the safety of the tuned models.
}
    \label{fig:random_tokens}
\end{figure}

Our investigation extended to using randomly selected tokens for regularization, with results presented in Fig.~\ref{fig:random_tokens}. We found that random tokens can also contribute to improved model safety. This effect can be explained through the lens of continual learning regularization, where preserving token-level information from the pretrained model may help maintain safety properties. However, the mechanism behind random tokens' effectiveness remains uncertain, as these tokens lack explicit connections to safety concepts. Still, this finding points to a broader principle: when aiming to preserve specific model behaviors, one can identify relevant tokens and apply token regularization to maintain desired characteristics.

\subsection{The trade-off between safety and targeted model adaptation}
Prior safety research has largely neglected a critical consideration: model performance on the target training data (evaluation loss). While existing safety enhancement methods improve alignment, they often sacrifice learning effectiveness on the original task. As demonstrated in Table~\ref{tab:safe_eval}, our method not only improves safety and utility but also achieves performance comparable to standard LoRA on the target data. These results demonstrate that our approach successfully balances safety requirements with task performance, showing that enhanced safety does not necessitate compromised learning capabilities on the core task data.

\begin{table}
\begin{center}
\begin{tabular}{ c|c|c|c}
&Tokens&ASR(\%) &eval\_loss(\(\downarrow\)) \\
\hline
&LoRA &16.7 &0.79\\
\hline
\multirow{2}{*}{$\lambda=1$}&I & 2.9&0.80\\
&cannot & 3.1&0.80\\
\hline
\multirow{2}{*}{$\lambda=2$}
&I & 0.58&0.81\\
&cannot & 1.9&0.81\\
\hline
\end{tabular}
\end{center}
\caption{\textbf{Trade-Off Between Safety and Targeted Data Performance.} The evaluation loss for both the traditional LoRA approach and our proposed method remains comparable. Increasing the importance of the token-loss term effectively suppresses harmful responses without substantially affecting the evaluation loss.
}
\label{tab:safe_eval}
\end{table}

\subsection{Analysis of running time}
Table~\ref{tab:running_time} compares the per-iteration running times of LoRA, DoRA, and our proposed method on the Commonsense-15k dataset using LLaMA-2-7b-chat, with a batch size of 16 and a LoRA rank of 32. Our method runs 1.34 times slower than LoRA but remains 1.25 times faster than DoRA.
\begin{table}
\begin{center}
\begin{tabular}{ c|c}
\hline
Method&Running time (ms)\\
\hline
LoRA & 435 (ms)\\
DoRA & 725 (ms)\\
Ours & 581 (ms)\\
\hline
\end{tabular}
\end{center}
\caption{The running time (in milliseconds) of our method compared with LoRA and DoRA.}
\label{tab:running_time}
\end{table}

\section{Conclusion}

We have revisited the critical challenge of safety preserving in LLMs during fine-tuning scenarios—a concern of increasing importance as these models are widely adapted for sensitive domains. We introduced safety token regularization (STR), a lightweight and readily implementable approach that leverages the inherent safety knowledge encoded within pre-trained models. Our extensive empirical evaluation across diverse benchmarks demonstrates that STR not only effectively preserves pre-trained safety, achieving state-of-the-art safety performance, but also maintains competitive task utility and, surprisingly, enhances training stability. By constraining the logits of salient safety tokens identified from rejection templates, STR offers a practical and readily deployable strategy for continual safety alignment in fine-tuned LLMs, filling a critical gap in current parameter-efficient fine-tuning methodologies. 

\paragraph{Current limitation and future works}
Despite demonstrating robust performance, our approach has some limitations that warrant further investigation. A key constraint is that our method restricts fine-tuned models to inherit the safety behavior of their pretrained counterparts, potentially limiting flexibility for new safety requirements. Although certain results suggest that our method can learn beyond the pretrained model’s safety scope, direct model regularization remains necessary. Moving forward, we plan to strengthen safety by both preserving existing knowledge and collecting or abstracting new insights from upcoming data.



\bibliography{custom}




\end{document}